\documentclass[letterpaper, 10 pt, conference]{ieeeconf}  %

\usepackage{algorithm}
\usepackage{algorithmic}
\usepackage{amssymb}		%
\usepackage{graphicx}		%
\usepackage{hypernat}		%
\usepackage{amsmath}
\usepackage{comment}

\usepackage{booktabs}
\usepackage{multirow}
\usepackage{graphicx}
\usepackage{bm}
\usepackage[backend=biber,style=ieee,natbib=true]{biblatex} 

\usepackage{etoolbox}
\makeatletter
\patchcmd{\@makecaption}
  {\scshape}
  {}
  {}
  {}
\makeatother

\IEEEoverridecommandlockouts                              %

\usepackage{amsmath} %
\usepackage{amssymb}  %
\usepackage[normalem]{ulem}
\usepackage{color}

\usepackage[utf8]{inputenc}
\usepackage{pgfplots}
\DeclareUnicodeCharacter{2212}{−}
\usepgfplotslibrary{groupplots,dateplot}
\usetikzlibrary{patterns,shapes.arrows}
\pgfplotsset{compat=newest}

\title{\LARGE \bf
Better Together: Online Probabilistic Clique Change Detection in 3D Landmark-Based Maps
}

\author{Samuel Bateman, Kyle Harlow, and Christoffer Heckman$^{\ast}$%
\thanks{All authors are with the Department of Computer Science, University of Colorado,
        Boulder, CO 80309, USA.}%
\thanks{$^{\ast}$Corresponding author. E-mail: \texttt{ christoffer.heckman @colorado.edu}.}%
}

\newcommand\copyrighttext{%
  \footnotesize \textcopyright 2020 IEEE.  Personal use of this material is permitted.  Permission from IEEE must be obtained for all other uses, in any current or future media, including reprinting/republishing this material for advertising or promotional purposes, creating new collective works, for resale or redistribution to servers or lists, or reuse of any copyrighted component of this work in other works.
  } %
\newcommand\copyrightnotice{%
\begin{tikzpicture}[remember picture,overlay]
\node[anchor=south,yshift=10pt] at (current page.south) {\fbox{\parbox{\dimexpr\textwidth-\fboxsep-\fboxrule\relax}{\copyrighttext}}};
\end{tikzpicture}%
}

\begin{document}

\maketitle
\copyrightnotice
\thispagestyle{empty}
\pagestyle{empty}

\begin{abstract}

Many modern simultaneous localization and mapping (SLAM) techniques rely on sparse landmark-based maps due to their real-time performance. However, these techniques frequently assert that these landmarks are fixed in position over time, known as the \emph{static-world assumption}. This is rarely, if ever, the case in most real-world environments. Even worse, over long deployments, robots are bound to observe traditionally static landmarks change, for example when an autonomous vehicle encounters a construction zone. This work addresses this challenge, accounting for  changes in complex three-dimensional environments with the creation of a probabilistic filter that operates on the features that give rise to landmarks. To accomplish this, landmarks are clustered into cliques and a filter is developed to estimate their persistence jointly among observations of the landmarks in a clique. This filter uses estimated spatial-temporal priors of geometric objects, allowing for dynamic and semi-static objects to be removed from a formally static map. The proposed algorithm is validated in a 3D simulated environment. 

\end{abstract}

\section{Introduction}\label{sec:intro}
As robots are deployed into real-world environments more often and for longer durations, it becomes increasingly important for them to be able to navigate in dynamic environments. 
Frequently, this is performed by providing spatial awareness through perception techniques such as simultaneous localization and mapping (SLAM), wherein map parameters are jointly estimated with robot pose in an online probabilistic framework. %
While SLAM is a very powerful tool for inference over core quantities for robotic perception, it %
can yield biased results due to %
core assumptions made, such as that the landmarks that compose the map are static in time. These assumptions are present at the level of the formalism employed to represent the problem. %
This work focuses on the limitations of this so-called \emph{static-world assumption} and how to extend the formalisms and algorithms involved in solving the SLAM problem in dynamic environments.

\begin{figure}[t]
\label{fig:oxford}
\begin{center}
    \includegraphics[width=0.48\textwidth]{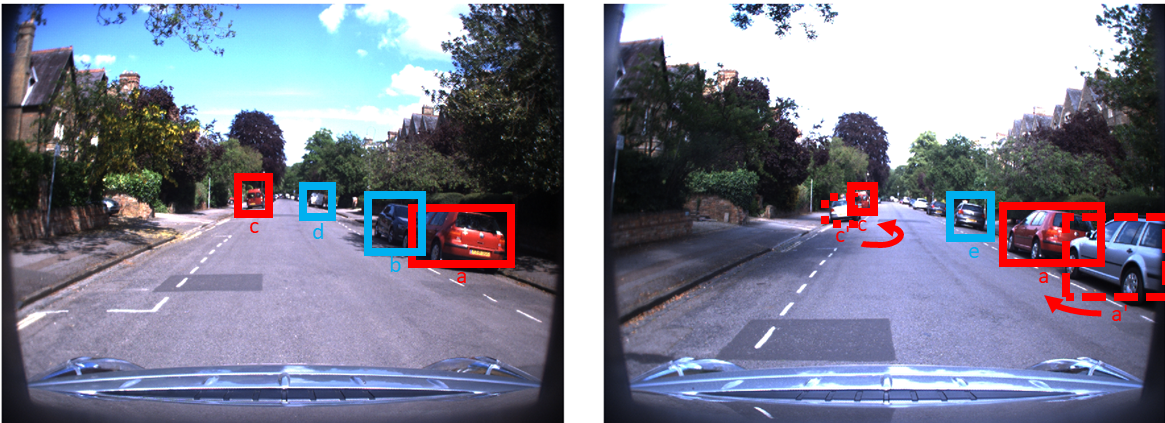}
\end{center}
\caption{Two images taken from the Oxford RobotCar dataset \cite{oxford} at similar locations approximately one month apart have drastic differences in appearance. Landmarks $a$ and $c$ have both shifted to $a'$ and $c'$ respectively, with new vehicles parked in their previous locations. Shifts in these landmarks from associating new measurements improperly could cause a localization system to estimate that it is in a different position. In addition, landmarks $b$ and $d$ have disappeared, while landmark $e$ has been added.}
\end{figure}

In dynamic-world deployments, such as those in warehouses, buildings, and on streets, %
one of the challenges that can lead to catastrophic failure in SLAM is the problem of associating features derived from data to landmark quantities to be estimated, frequently termed the ``data association problem.'' %
Data association matches a measurement from a sensor to a state parameter to be estimated, e.g.\ a visual feature such as a corner in an image might be associated with the corner of an office chair \cite{SIFT,SURF,ORB,DEEPFEATURES,JCBB,Rosen,Cooper}. %
However, if that chair moves for any reason, the measurement of the same corner may be associated with the previous landmark %
position, rather than identifying that %
the observed feature %
should be %
used to generate %
a new landmark %
independent of any landmarks already in the map. %
Without properly identifying these associations, the SLAM solution will %
at best incur error, and at worst diverge entirely \cite{tracking}. %
This is particularly problematic in long-term deployments, in which landmarks in the map are likely to vary in position and possibly even appearance over time. %
A pertinent example from the Oxford RobotCar Dataset \cite{oxford} is shown in Figure \ref{fig:oxford}. 
While there are a number of potential data association techniques \cite{JCBB, MLDA},  %
most do not account for dynamic environments, leading to almost all SLAM approaches making a static world assumption.  %
While this assumption holds for short deployments in controlled environments, it is clear that long-term real-world deployments will violate this assumption. 

This work attempts to account for dynamic maps by applying a Bayesian filter over the persistence of map parameters at the core of modern SLAM implementations, inspired by previous persistence filters developed by \cite{Nobre,Rosen}. 
Further, this work demonstrates the advantages of conditioning over arbitrary groupings of highly correlated landmarks as well as the objects they are associated to while accounting for differing sensing modalities and the degradation of said sensing modalities, which is unique to this work relative to the previous work in Bayesian landmark persistence filters such as \cite{Nobre, Rosen}.

\section{Prior Work}\label{sec:prior_work}
The problem of deploying robots in dynamic environments has been explored since the formulation of SLAM \cite{BayesDynamics}. %
The primary challenges of dynamic SLAM, detecting changes and updating the map, have been addressed several ways. %
Early approaches, using the extended Kalman filter \cite{Smith1990} and particle filter SLAM \cite{FastSLAM}, focused on separating the map into static and dynamic portions, %
thereby reducing %
the problem back to a static SLAM problem \cite{Wolf2005}. 
Other approaches focused on tracking the dynamic objects using a separate algorithm such as a human vision model \cite{Weikersdorfer2013Simultaneous}. 
In sparse landmark-based SLAM, %
early work focused on updating maps hierarchically \cite{hierarchicalObject}. Others focused on randomly selecting portions of the map to revisit throughout a run to update the prior map \cite{biber2009experimental,Biber2005DynamicMF}. Another set of proposed solutions involved increasing the storage of the map, such as in \cite{multimap} where a multiple map approach was developed. 
The goal of these approaches was to store multiple maps and select the map that best fits the current sensor measurements. 
Another multi-map approach stored experiences tied to specific places, and then selected the best possible set of experiences for navigation \cite{experience}. 
While this work was robust to changes in lighting, weather, and seasonal changes, it suffered in areas where the experience was consistently changing, such as in parking lots. 

Many other approaches focus on specific sensor modalities: 
\cite{Tan} focused primarily on monocular SLAM for dynamic environments, while \cite{Zou} used a similar approach including multiple cameras with dynamic priors assigned to features tracked by each camera. 
Additional studies have focused primarily on laser scanners, but relied on other sensors to help reduce pose error in the presence of dynamic objects \cite{Zhao}. 
Many of these sensor-based approaches lack the adaptability of an approach that is based purely on the parameters present in the SLAM formulation. 

Some more recent approaches have begun to address these problems by tracking dynamic features explicitly or jointly estimating objects' motion relative to the current platform \cite{8660984,8460681}. 
In our work, we focus instead on filtering the features based on an initial prior accounting for the 3D semantic spatial relationships between collections of correlated features. 
To do this, we propose a new approach in the spirit of \cite{Rosen,Nobre} which tracks the joint persistence probability of one of these collections of landmarks and removes them from the factor graph when the posterior probability reaches a tunable threshold \cite{thrun2006graph}.

\section{Methodology}\label{sec:methodology}

Formulating map change detection in relation to a probabilistic SLAM problem formulation begins with introducing the SLAM problem.
In its simplest form, SLAM gives the robots' pose $\bm{x}_p \in \mathbb{SE}(3)$ and map landmarks $\bm{x}_l \in \mathbb{R}^3$ based on measurements $Z$ which can include odometry information from the robot as well as landmark detections, defined by a sensor model. 
With this information, the goal of maximum \textit{a posteriori} (MAP) estimate SLAM is to maximize the posterior $p(\mathcal{X}|Z)$ where $\mathcal{X}=[\bm{x}_p, \bm{x}_l]$. 
While this is possible given a limited number of measurements, with any reasonably-sized map this would require summing over all possible measurement associations to obtain the full probability distribution, leading to an intractable calculation \cite{Dellaert-2001-8303}. 
As such, in most problems possible measurement associations are restricted through a data association step as mentioned in Section \ref{sec:intro}. 
This is done by forming a vector of data association hypotheses $J$. 
Ideally, this vector is developed by considering the most likely data associations given our measurements $Z$ and calculating $\text{argmax}_J P(J|Z)$. 
However, matching a sequence of measurements to a series of discrete landmarks, while also accounting for spurious measurements, is a combinatorially hard problem \cite{JCBB}.  
As such, most SLAM algorithms instead use a known data association technique such as joint compatibility branch-and-bound (JCBB) \cite{JCBB} or maximum likelihood data association \cite{MLDA} based on sensor model-derived landmark similarity metrics. 
For a survey of data association techniques, see \cite{Cooper}.  
Thus, the typical SLAM problem with the static-world assumption is $\text{argmax}_{\mathcal{X}}P(\mathcal{X}|J,Z)$ where $J$ is now the vector of data associations assigning a single measurement to a single landmark. %
This formulation is extended by the introduction of landmark cliques, denoted $\tau$, any one of which is a collection of correlated landmarks, e.g.\ those on the surface of a rigid body.

A similar approach was taken in \cite{Nobre}, however the formulation is expanded to introduce the usage of arbitrary, rigid cliques to the previous methodology. 
The persistence of a clique is represented as a time-varying, binary random variable $\theta_{\tau}^{t}\in\{0,1\}$ where $t\in [0, \infty)$ and $t=0$ is the first time the clique was observed. %
Ideally, a joint calculation of the feature association hypothesis $J$ and the existence of a landmark in a clique $\theta^t$ given a set of measurements $Z$ would be possible, but the same problems of intractability and spurious measurements calculating $P(\theta^t, J|Z)$ exist as they would in calculating $P(J|Z)$. 
Instead, the landmark persistence is conditioned on the data associations, calculating $\text{argmax}_{\theta^t}P(\theta^t|J)$ by estimating $J$ using a known data association technique and then estimating $\theta^t$.
From these results, the SLAM algorithm decides whether or not to continue integrating measurements of a particular element of $J$ based on the probability of $\theta_\tau$, or to introduce a new feature into the map, while removing the old landmark, similar in spirit to the ideas of \cite{Nobre} and \cite{Rosen}. 
In this way, $\mathcal{X}$ is augmented as $\mathcal{X} = [x_p,x_l,\theta^t]$ and jointly estimated in $\text{argmax}_{\mathcal{X}}P(\mathcal{X}|J,Z)$ as before.\

\subsection{Clique Persistence}

In \cite{Rosen,Nobre}, it was assumed that the $k^\text{th}$ feature in a map $M$ has a survival time $T_k\in [0, \infty)$ such that the feature should be expected to no longer persist for $t > T_k$.
This connection between survival time and persistence can be formalized as
\begin{equation}
\label{old:persist}
P(\theta_k^t | J) = P(T_k \geq t_n | J_{k}^{1:N}),
\end{equation}
where $j^{t_i}_k$ is the random variable describing the state of a detection of a feature $k \in M$ at possible observation time $t_i$, denoting the sequence of observations of $k$ as
\begin{equation}
J_k^{1:N} \triangleq \{j_k^{t_i}\}_{i=1}^N.
\end{equation}
However, the formulation presented in these works does not account for the fact that, in 3D, most features on a given object are not observable a large part of the time, even when they are within our perceptive field.
The clique survival time

\begin{equation}T_\tau \sim P_{T_\tau}(\cdot), \label{eq:clique_persistance}\end{equation}

\noindent is used to address this, where $P_{T_\tau}(\cdot)$ is approximated by a prior distribution over the survival time $T_\tau$, similar to the feature survival time in \cite{Rosen}. 
To account for the partial observability of landmarks, a similar relation is made to the clique survival time $T_\tau$ and the previously introduced clique persistence random variable $\theta_\tau^t$ as was done in Eq.\ \eqref{old:persist}; that is,
\begin{equation}
\label{new:persist}
P(\theta_\tau^t | J) = P(T_\tau \geq t_n | J_{\tau}^{1:N}),
\end{equation}
where $J_\tau^{1:N}$ is the shorthand for $J_{k\in\tau}^{1:N}$.
This carries with it the assumption that $\forall i \in \tau, T_i = T_\tau$, which simply states that all landmarks within a clique are likely to persist so long as the clique persists.
This deceptively small change has a major implication: an individual landmark may be observed for only a few frames of observation, but some non-empty subset of the clique of landmarks will almost always be observable at every timestep of observation. 
This also assumes that clique persistence is independent of the observations of other cliques.
This assumption of independence is justified by the observation that clique survival time is modeled to be independent of each other clique survival time.
Since observability of any given landmark is strongly correlated with its persistence (precluding missed and false detections which are addressed later), it seems reasonable that the observation of landmarks in other cliques would provide little extra information about clique persistence then can be provided by only the sequence of observations of the component landmarks of a given clique.
While this may not be strictly true in an analytical sense, it is a necessary assumption to maintain tractability of the presented problem, as also identified by \cite{Nobre}.

Using Eq.\ \eqref{new:persist}, Bayes' Rule is then employed to estimate the persistence of a clique at a given time $t$
\begin{equation}\label{eq:bayes}
P(T_\tau \geq t| J_\tau^{1:N}) = \frac{P(J_\tau^{1:N} | T_\tau \geq t_n) P(T_\tau \geq t)}{P(J_\tau^{1:N})}.
\end{equation}
From here, all that remains is to estimate the likelihood and the evidence of the clique to estimate persistence.

\subsection{Joint Detection Likelihood}
In order to calculate the joint likelihood of observations of a clique, an assumption that, for $a, b \in \tau$
\begin{equation}
\begin{split}
    P(J_a^{1:N}, J_b^{1:N} | T_\tau \geq t_n)& = \\
    \quad &P(J_a^{1:N} | T_\tau \geq t_n) P(J_b^{1:N}| T_\tau \geq t_n)
    \end{split}
\end{equation}
is made, which is to say that a sequence of detections for one landmark $a$ is conditionally independent from each other landmark e.g. $b$, in the clique, given the persistence of the clique.
With this, the joint clique likelihood can be broken into the product of detection likelihoods of individual landmarks

\begin{equation}\label{eq:likelihood}
P(J_\tau^{1:N} | T_\tau \geq t_n) = \prod_{k \in \tau} P(J_k^{1:N} | T_\tau \geq t_n).
\end{equation}
Another assumption is that, given the persistence of a feature $k\in \tau$, the sequence of detections will be fully Markovian, as we do not incorporate information about the projected trajectory of the sensor.
The definition of landmark likelihood follows as:
\begin{equation}
    P(J_k^{1:N} | T_\tau \geq t_n) = \prod_{i=1}^N P(j_k^{t_i} | T_\tau \geq t_n),
\end{equation}

\noindent which then leads to the fully expanded joint detection likelihood
\begin{equation}
\begin{split}
    P(J_\tau^{1:N}|T_\tau\geq t_i) &= \prod_{k\in\tau} P(J_k^{1:N}|T_\tau\geq t_i)\\
    &= \prod_{k\in\tau} \prod_{i=1}^N P(j_k^{t_i}|T_\tau\geq t).
\end{split}
\end{equation}

\noindent By identifying that the $t_{N}$ likelihood is in the expression for the $t_{N+1}$ likelihood, the $t_{N}$ expression can be factored out. The clique likelihood can be defined recursively as
\begin{equation}\label{eq:rec_likelihood}
    P(J_\tau^{1:N+1}|T_\tau\geq t) = P(J_\tau^{1:N}|T_\tau\geq t_i)\prod_{k\in\tau} P(j_k^{N+1}|T_\tau\geq t),
\end{equation}
requiring only the new measurements acquired at time $t_{N+1}$ to update the clique likelihood.

Finally, the individual detection likelihoods $P(j_k|T_\tau)$ must be calculated in order to fully define the clique likelihood in Eq.\ \eqref{eq:bayes}. 
With a perfect sensor, this would simply be a binary random variable such that
\begin{equation}
\label{eq:perfect_detection}
    P(j_k|T_\tau) = \begin{cases}
    1, & T_\tau \geq t\\
    0, & T_\tau < t,
    \end{cases}
\end{equation}
in the case where $\tau$ is expected to be observable by the sensor. This does not however account for the probability of falsely detecting a non-existent landmark, denoted $P_F$, or the probability of missing detection for an existing landmark, denoted $P_M$, which is particularly troublesome in 3D due to occlusion and sensor degradation, e.g. sensor range limitations. 
Previous works utilized a fixed constant for $P_F$ and $P_M$, but this does not capture the complexity of real sensors in general.
To remedy this, sensor degradation is modeled by making $P_M$ a function of minimum clique landmark distance from the sensor, denoted $P_M(s)$. 
It is common for perception systems to have a maximum detection range $s_\text{max}$ but for measurements to only be incorporated into a model, such as a SLAM system, up to a smaller threshold $s_\text{obs}$ where sensor data is more reliable.
This notion is used to define a range-based sensor degradation model $P_M(s)$ as follows:
\begin{equation}\label{eq:range_prior}
    P_M(s) = 1-\exp{(-s\cdot s_\text{max}/s_\text{obs})}.
\end{equation}
\noindent With $P_M$ and $P_F$ defined, the formulation of the feature detection likelihood from \cite{Rosen}
\begin{equation}
\label{eq:detection}
    P(j_k|T_\tau) = \begin{cases}
    P_M^{1-j_k^t}(1-P_M)^{j_k^t}, & T_\tau \geq t\\
    P_F^{j_k^t}(1-P_F)^{1-j_k^t}, & T_\tau < t
    \end{cases}
\end{equation}

\noindent is used in this filter.

\subsection{Joint Evidence}
The derivation of the joint clique evidence is inspired by \cite{Rosen}.
This work capitalizes on the conditional independence of landmark detection within a clique given clique persistence.
Because the clique likelihood defined in Eq.\ \eqref{eq:likelihood} is only updated at discrete times $\{t_i\}_{i=1}^N$, the likelihood is constant over the intervals in the set
\begin{equation}
    \{[t_i, t_{i+1}) | \forall i \in \mathbb{Z}, 0 \leq i \leq N\}.
\end{equation}
\noindent As in \cite{Rosen}, the integral is simplified by defining $t_0\triangleq0$ and $t_{N+1}\triangleq\infty$.
Thus the evidence over cliques is calculated as
\begin{equation}
\begin{split}
P(J_\tau^{1:N})&=\int_0^\infty P(J_\tau|T_\tau)P(T_\tau)dT_\tau\\     
&=\sum_{i=0}^N\int_{t_i}^{t_{i+1}}P(J_\tau|T_\tau)p(T_\tau)dT_\tau\\
&=\sum_{i=0}^N\prod_{k\in\tau}P(J_k^{1:N}|T_\tau)[F_{T_\tau}(t_{N+1}) - F_{T_\tau}(t_N)],
\end{split}
\end{equation}
where $F_{T_\tau}(t)$ is the cumulative distribution function of $P_{T_\tau}(t)$ which is described in Eq.\ \eqref{eq:clique_persistance}. In the spirit of \cite{Rosen}, the clique evidence $P(J_\tau^{1:N})$ is broken into a lower partial sum
\begin{equation}
    L(J_{\tau}^{1:N})\triangleq\sum_{i=0}^{N-1}\prod_{k \in \tau}P(J_k^{1:N}|T_\tau)[F_{T_\tau}(t_{N+1}) - F_{T_\tau}(t_N)],
\end{equation}

\noindent so that the evidence is defined recursively as
\begin{equation}\label{eq:evidence}
    P(J_\tau^{1:N}) = L(J_{\tau}^{1:N})+\big(\prod_{k \in \tau}P(J_k^{1:N}|T_\tau)\big)[1 - F_{T_\tau}(t_N)].
\end{equation}

\noindent Then $L(J_{\tau}^{1:N})$ is related to $L(J_\tau^{1:N+1})$ as
\begin{equation}\label{eq:partial_clique_evidence}
\begin{split}
    L(J_\tau^{1:N+1}) &= \big(\prod_{k \in \tau} P_F^{j_k^{t_{N+1}}} (1 - P_F^{j_k^{t_{N+1}}})\big)\\
    \big(L(J_{\tau}^{1:N}) &+ P(J_\tau^{1:N} | t_n \geq T_\tau)[F_T(t_{N+1}) - F_T(t_N)]\big),
\end{split}
\end{equation}

\noindent admitting the full recursive Bayesian estimation procedure as outlined in Algorithm \ref{alg:estimation}. 
\begin{algorithm}
\caption{Algorithm for recursively calculating joint clique filter (JCF) posterior persistence probability}
\label{alg:estimation}
\begin{algorithmic}[1]
\renewcommand{\algorithmicrequire}{\textbf{Input:}}
\renewcommand{\algorithmicensure}{\textbf{Output:}}
\REQUIRE  Feature detector error probabilities ($P_M$, $P_F$), cumulative clique prior $F_{T_\tau}$, feature detector outputs $J_{\tau}^{t_i}$.
\ENSURE Persistence beliefs on clique $P(\theta_\tau = 1|J_\tau^{1:N})$ for $t\in[t_N,\infty)$.
\\ \textbf{Initialization:} Set $t_0\leftarrow 0$, $N\leftarrow 0$, $P(J_\tau^{1:0}|t_0)\leftarrow 1)$, $L(J_\tau^{1:N})\leftarrow 0, P(J_\tau^{1:N})\leftarrow1$
\WHILE{$\exists$ new data $J_\tau^{t_{N+1}}$}
\STATE Compute the partial clique evidence $L(J_\tau^{1:N+1})$ using \eqref{eq:partial_clique_evidence}
\STATE Compute the clique likelihood $P(J_\tau^{1:N+1}|T_\tau\geq t)$ using \eqref{eq:rec_likelihood}
\STATE Compute the clique evidence $P(J_\tau^{1:N+1})$ using \eqref{eq:evidence}
\STATE $N\leftarrow N+1$
\STATE Compute the posterior probability $P(T_\tau \geq t|J_\tau^{1:N})$ using \eqref{eq:bayes}
\ENDWHILE
\end{algorithmic}
\end{algorithm}

\subsection{False Negative Suppression}
A major shortcoming of Bayesian landmark filters in 3D is that cliques, and their constituent landmarks, are only partially observable due to sensor degradation, map geometry, and map occlusion.
However, in a joint filter formulation such as in \cite{Nobre} and here, the filter must be updated for all landmarks in a clique if any one of the constituent landmarks is visible.
This works well when the clique is well within sensor range, but becomes problematic when only a few of the landmarks within a clique are possibly observable, such as when the clique is on the boundary of the observable area.
The filter would then update, including in a number of negative detections for which there is no corresponding real information, driving the persistence probability down prematurely.

To remedy this, for all landmark filters studied in this work, a technique to suppress false negatives is introduced: 
Let $d(\bm{x}_p, \bm{x}_{l_k})$ be defined as the Euclidean distance between the positional component of $\bm{x}_p$ and $\bm{x}_{l_k}$.
Let 
\begin{equation}
\tau^* = \{k \in \tau | k \in \text{positive detections at }t_n\}
\end{equation}
and pick $\delta$ to be a positive threshold ratio of positive detections to total expected detections.
This parameter would likely be tuned for the application, however for an example and purposes of simulations, our $\delta$ was chosen arbitrarily to be $0.03$.
If the following 3 statements are true

\begin{align*}
    1. \quad & \forall k \in \tau, s_\text{obs} < d(\bm{x}_p, \bm{x}_{l_k}),\\
    2. \quad & \exists k \in \tau \text{ st. } d(\bm{x}_p, \bm{x}_{l_k}) <  s_\text{max}, \text{and}\\
    3. \quad &\frac{|\tau^*|}{|\tau|} < \delta,
\end{align*}

\noindent then reject detections for all of $\tau$ at this time step. While deceptively simple, this can significantly improve the performance of the joint clique filter in 3D, as well as the joint feature filter from \cite{Nobre} and even the independent feature filter from \cite{Rosen}, which eschews any inter-feature correlation model.

\section{Experiments and Results}\label{sec:experiments_and_results}

\begin{table*}[hbtp]
\vspace{10mm}
\centering
\begin{tabular}{@{}l|l|llll|llll@{}}
\multirow{2}{*}{Sensor Model} & \multirow{2}{*}{Metric} & \multicolumn{4}{l|}{No Negative Suppression}    & \multicolumn{4}{l}{Negative Supression}                 \\ 
                              &                         & IFF            & JFF   & JCF    & JCFR           & IFF            & JFF   & JCF            & JCFR           \\ \midrule
\multirow{2}{*}{Lidar}        & Accuracy                & 0.715          & 0.621 & 0.850 & \textbf{0.942} & 0.762          & 0.618 & 0.978          & \textbf{0.987} \\
                              & MES/S                   & 0.144          & 0.099 & 0.099 & \textbf{0.308} & 0.211          & 0.012 & 0.479          & \textbf{0.788} \\ \midrule
\multirow{2}{*}{Camera}       & Accuracy                & 0.672          & 0.587 & 0.793 & \textbf{0.844} & 0.731          & 0.603 & \textbf{0.936} & 0.916          \\
                              & MES/S                   & \textbf{0.254} & 0.025 & 0.179 & 0.170          & \textbf{0.336} & 0.037 & 0.326          & 0.300         
\end{tabular}%
\caption{IFF: Independent Feature Filter \cite{Rosen}, JFF: Joint Feature Filter \cite{Nobre}, JCF: Joint Clique Filter (Ours), JCFR: Joint Clique Filter with range sensor degradation (Ours). MES/S is the mean estimated survival time over true survival time which compares the rate at which landmarks were removed to when they should have been removed (Higher is Better). Each of the tested filters attempts to filter changed landmarks from the factor graph. When comparing the JCF/JCFR with the IFF, it is of note that the IFF can do arbitrarily badly depending on the environmental occlusion of landmarks, whereas the JCF and JCFR integrate a number of landmark measurements from a single clique at each time step and are therefore is much more robust to occlusion in 3D environments. For the purposes of fair evaluation, this problem of the IFF is not explored in these results. The JFF can be directly compared to our methods. }

\label{tab:compare}
\end{table*}
\subsection{Simulations}
\begin{figure}[tbh]
\centering
\resizebox{0.48\textwidth}{!}{\input{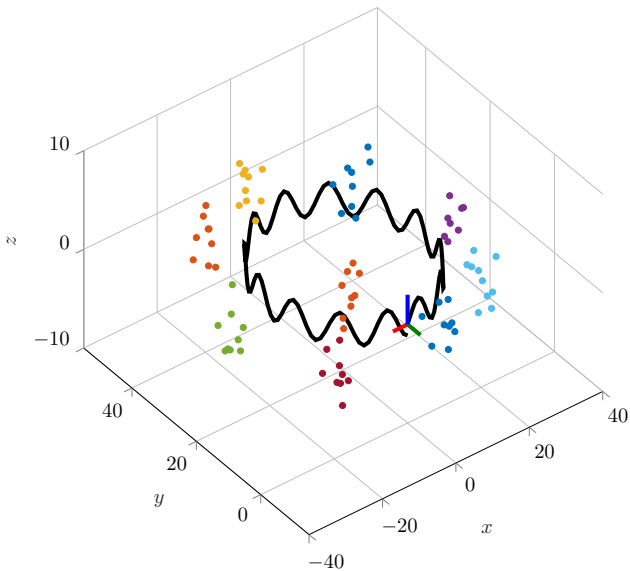}}
\caption{Example single-period sensor trajectory simulation with all cliques visible. Most landmarks would not be visible to either of the sensor modalities at the shown time step due to their orientation and the limited field of view of the sensors.} %
\label{fig:sim}
\end{figure}
To accurately test the relevant algorithms on realistic 3D data, a 3D simulation of arbitrary, periodic 3D trajectories for a sensor moving past a collection of 3D objects with oriented features corresponding to landmarks, with normals equal to the vector from the center of the object to the feature point was developed.
The simulation models two sensor types: a monocular camera and a 3D lidar.
Both sensors have a canonical forward orientation but do not generally move in the forward direction.
In simulation we first define $s_\text{max}$, the range within which the sensor can detect features, and model detection probability as $P(k) = \exp{(d(\bm{x}_p, \bm{x}_{l_k})/(\lambda))}$ with $\lambda = 0.7 \cdot s_{\text{max}}$.
For the camera, we use a pinhole projection model where object detection candidates are decided by being within the field of view of the camera oriented along the local $x$-axis. 
We model the lidar returns along an annulus around the vertically-aligned coordinates of the sensor frame with a field of view angle above and below the $xy$ plane of the sensor frame.

In the simulations, each object has a uniformly-chosen random number $U\sim[15,25]$ of features corresponding to landmarks on their surface.
Like real landmarks in 3D, corresponding features are not observable from all angles, due to occlusion by the object on which the landmark lies or due to the maximum viewing angle at which a landmark can be detected, distorting their feature-based appearance \cite{ORB}.
This is modeled in simulation by having a maximum angle from the normal of the object at that landmark which is approximated as the vector from the center of the object to the point.
If a sensor could detect a landmark that is outside this maximum angle of observation, it will only be detected with probability $P_F$.
The objects stop persisting at a certain time in the simulation, similar to \cite{Rosen}, to simulate the movement of a semi-static object or the change in a long term structure.

1000 simulations were run over arbitrary 3D periodic trajectories that pass by the same clique multiple times. 
An example trajectory and set of cliques are shown in Figure \ref{fig:sim}.
All filters are run over the same trajectories for each run to make it as fair as possible. We denote true positives as \textit{TP}, False Positives as \textit{FP}, True Negatives as \textit{TN}, False Negatives as \textit{FN} and the total number of possible detection time steps as \textit{TR}. 
Our accuracy metric is defined as is traditional as ${(\textit{TP} + \textit{TN})}/{\textit{TR}}$,
where a positive is defined as above and negative is defined as below a persistence probability threshold $\rho_h$ to allow a measurement to be incorporated to the factor graph.

\subsection{Results}

\begin{figure}[t]
\centering
\resizebox{0.48\textwidth}{!}{\input{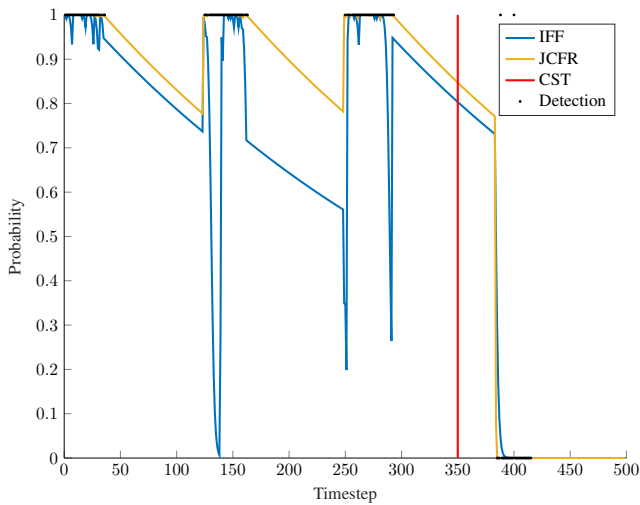}}
\caption{Comparison of example simulation of the independent feature filter (IFF) \cite{Rosen} against the joint clique filter with range sensor degradation (JCFR), plotting clique survival time (CST) and at least 1 detection (1) or no detections (0) detections from the clique for each expected observation (D). Note that the independent feature filter is far less stable and has many discrepancies with the joint clique filter with range sensor degradation due to the inherent partial observability of landmarks in 3D.}
\label{fig:time_series}
\end{figure}

During an example run, we compare various filters by plotting the posterior probability of a given landmark within a clique. 
In the SLAM problem, a filter will remove that landmark from the map when the posterior drops below a certain threshold $\rho_l$, and reject new measurements when the posterior probability drops below $\rho_h$. 
As shown in Figure \ref{fig:time_series}, the independent feature filter removes the given landmark several times, prior to its actual survival time. 
The joint clique filter with range sensor degradation on the other hand is able to reason over the various missed detections, only removing the landmark from the map after the first observations after the landmarks survival time passes at $t=350$.

The full results of our simulations can be seen in Table \ref{tab:compare}.
The joint clique filter outperforms the others in accuracy, which is quite important as the Precision of all filters is almost always greater than $99\%$, meaning that nearly all of the error is coming from false negatives for each of the filters. 
A high false negative count will significantly decrease the number of measurements that can be incorporated into the SLAM algorithm over time, thereby limiting the accuracy of the map and pose estimation, potentially quite significantly.
Based on the accuracy of our filters, even in a fully static environment, there should be little to no loss of accuracy while incorporating the ability to detect and remove semi-static and dynamic landmarks from a map.

We also define a metric called the mean estimated survival time over true survival time (MES/S) metric which, for a given clique $\tau$, is defined as
\begin{equation}\label{eq:mess}
     \text{MES/S} = \frac{1}{R} \sum_{r=0}^R \frac{\hat{T}_\tau^r}{T^r_\tau},
\end{equation} where $R$ is the number of runs in a set of simulations, $T_\tau^r$ is the actual realization of $T_\tau$ in the $r^\text{th}$ simulation and $\hat{T}_\tau^r$ is the time at which the filter removes the landmark from the map. 
This metric quantifies the efficiency of the landmark filters in a SLAM system as keeping landmarks in a map for the duration of their persistence in the real world. 
This is discussed in detail in Section \ref{sec:discussion}.
For lidar, our results represent an improvement over both the independent feature filter (IFF) \cite{Rosen} and the joint feature filter (JFF) \cite{Nobre}, where the joint clique filter (JCF) and joint clique filter with range sensor degradation (JCFR) achieve a relatively high MES/S with negative suppression, nearing a perfect score of 1.0, and therefore imposing little to no overhead than necessary over the SLAM algorithm to provide feature persistence detection.
For cameras, while our method still does well with negative suppression, it is much closer to the independent feature filter in terms of performance under ideal conditions for the IFF.
Our algorithm is not particularly designed for camera sensing modalities, in particular, the range-based sensor degradation and negative suppression are particularly well-tailored to the problems that occur in lidar sensing.
It is important to note, however, that in all of these sensing modalities the IFF can do arbitrarily badly on both the accuracy and MES/S metric given occlusions which commonly occur in 3D environments of interest. 
This makes a direct comparison of the methods difficult as the IFF lacks the inherent robustness of a joint filter utilizing all the information of the observed landmarks in the clique.
None the less, the JCF and JCFR still both perform competitively against the IFF in the Camera sensing modality in the MES/S metric, and better in every other metric and sensing modality against both the IFF and the JFF.

\section{Discussion}\label{sec:discussion}
When incorporated into a real SLAM pipeline, a negative detection would result in the removal of a landmark from the factor graph.
Thus, a filter which underestimates the true survival time extensively of a landmark would cause excessive operations to be performed on the factor graph and measurement information to be lost. 
For this reason, we introduced MES/S defined in Eq.\ \eqref{eq:mess}, which accounts for the inefficiency of a filter in prematurely removing landmarks from a graph before it has changed in reality. 
As the goal of these filters is to keep landmarks in the map for as long as possible without incurring error, this is a useful measure of how efficiently they perform the task they are designed to solve when combined with a high accuracy.
Ideally, this metric should be as close to 1 as possible, indicating that landmarks are rarely removed from the map prematurely and then added back in afterwards. 
The higher the value, the less early removals from the map and thus the smaller penalty for performance and accuracy of the SLAM algorithm.
A very low MES/S value would indicate constantly removing landmarks and subsequently all measurement constraints from the factor graph, significantly weakening the estimation.

This is a major weakness of the filters from the previous work \cite{Nobre,Rosen} we compare to here as they often will produce a false negative very early due to the geometric nature of the features which is not handled in either of these works, pushing this metric lower.
To allow these previous filters to perform better on the MFS/S metric, we defined a removal threshold $\rho_l$ which is less than our measurement filter threshold $\rho_h$ as the persistence probability threshold to remove a landmark from the map.

In addition, the tracked landmark in the clique for simulations of the independent feature filter \cite{Rosen} is always observable from the trajectory so that it does not unfairly get suppressed by trajectories that could not detect their corresponding landmark, but can still be undermined by random missed detections.
These modifications do give an unfair edge to the previous methods, however, we believe it is better to compare our filter in a more difficult scenario because it makes any positive results stand on their own.
In reality, the problem of the observability for the independent feature filter is much more dire when applied to a real SLAM system.
At every observable time step, nearly all occluded landmarks in a scene will be quickly removed from the map by the independent feature filter, vastly weakening the strength and reusability of the map for any trajectory except for the one being followed by the current sensor.
This severely limits the independent feature filter's practical applicability to 3D SLAM workloads, which will not be able to perform loop closure if a large number of landmarks have been removed from the map upon revisiting a location.
Our method addresses this major drawback, having the ability to keep landmarks that cannot be observed in the map while still removing dynamic and semi-static landmarks.

The addition of the sensor degradation range prior from the JCF to the JCFR on the LiDAR appears to be a strict improvement, modeling well the way that LiDAR detects features and observes objects.
This does not necessarily seem to be the case in the camera modality though.
This is likely because objects can get very close to the camera and then exit the frame, which would denote through the range prior that the filter should nearly always detect the landmarks, but the landmarks would suddenly disappear.
This also gets around the negative suppression, explaining why the JCF is dominant with a much smaller margin in this sensing modality, particularly with the JCFR.
It is likely, although we do not show, that a similar improvement in camera sensing modalities could be seen by adding a negative suppression based on viewing angle relative to the principle axis of the camera based on Field of View (FOV) to the negative suppression or sensor degredation model to handle the inherent boundary on the edge of the FOV which is similar to the range based boundary seen in lidar which was addressed with the presented negative suppression framework.

Finally, the false negative suppression improves the performance of all filters except for maybe the JFF from \cite{Nobre}, which doesn't see much performance improvement from the method.
It identifies a major problem of these filters: the fact that all potentially observable landmarks must be updated at once and landmarks are poorly observable at larger distances and angles of observation. 
False negative suppression provides a simple and satisfying solution to this problem.

\section{Conclusions}\label{sec:conclusions}
In this paper, we proposed a clique-based change detection model to account for non-static features in feature based SLAM algorithms. 
This work expands on previous work developing probabilistic change detection filters, which primarily focused on individual \cite{Rosen} or joint features \cite{Nobre}. 
While these works provide robust frameworks in 2D environments, they both have limited applicability in 3D where determining individual feature survival times is more difficult due to constant occlusions, less direct sensor models, and reliance on highly robust features. 
By filtering over cliques we showed our algorithm is more robust to these types of errors achieving both better accuracy and comparable or better MES/S making this a more efficient and robust algorithm. 
In addition, it is able to account for correlations between features in a simpler manner than \cite{Nobre}, by modelling entire cliques jointly.

\section*{Acknowledgment}
This work was supported by the National Science Foundation National Robotics Initiative 2.0 project \#1830686 ``Life-Long Learning for Motion Planning by Robots in Human Populated Environments'' as well as the Defense Advanced Research Projects Agency Subterranean Challenge ``MARBLE.''

\printbibliography

\end{document}